\documentclass[sigconf]{acmart}

\usepackage{atbegshi}
\AtBeginDocument{\AtBeginShipoutNext{\AtBeginShipoutDiscard}}

\usepackage{booktabs} 

\usepackage{etoolbox}
\makeatletter
\patchcmd\@combinedblfloats{\box\@outputbox}{\unvbox\@outputbox}{}{%
   \errmessage{\noexpand\@combinedblfloats could not be patched}%
}%
 \makeatother

\usepackage{pdfsync}
\usepackage{amsmath}
\usepackage{array}
\usepackage[caption=false,font=footnotesize]{subfig}
\usepackage{url}
\usepackage{float}
\usepackage{hyperref}
\usepackage{comment}
\usepackage{tikz}
\usepackage{pgfplots}
\pgfplotsset{compat=1.14}
\usepackage{pgfplotstable}
\pgfplotsset{select coords between index/.style 2 args={
    x filter/.code={
        \ifnum\coordindex<#1\fi
        \ifnum\coordindex>#2\fi
    }
}}
\usetikzlibrary{calc,positioning,shapes,shadows,arrows,arrows.meta,trees,intersections,patterns,plotmarks}
\usepackage{xparse}

\let\Intersect=\Intersection
\let\In=\in
\newcommand{\comma}{\ifmmode\text{,}\else\textrm{,}\fi}

\def\Luminance{Y}

\def\Sharpness{S}

\def\Uniformity{U}

\def\Red{R}
\def\Green{G}
\def\Blue{B}

\newcommand{\Gradient}[1]{G_{#1}}
\def\Histogram{H}

\def\FOneMeasure{F$_\text{1}$-measure}

\newcommand{\Video}[1][]{V_{#1}}

\newcommand{\Features}[2]{X^{#1}_{#2}}

\ExplSyntaxOn
\DeclareDocumentCommand{\Frame}{o d()}
{
  f\IfValueTF{#1}{\sb{#1}}{}\IfValueTF{#2}{^{#2}}{}
}
\ExplSyntaxOff

\newcommand{\Ranking}[1][]{R_{#1}}

\newcommand{\Segment}[1][]{s_{#1}}
\newcommand{\Segmentation}[1][]{S_{#1}}
\newcommand{\Summary}[1][]{U_{#1}}

\newcommand{\Duration}[1]{d_{\text{#1}}}
\def\TargetDuration{W}
\def\Weight{w}
\newcommand{\Channel}[1]{}

\def\UserSummaries{J}
\newcommand{\Precision}[1]{p_{#1}}
\newcommand{\Recall}[1]{r_{#1}}
\newcommand{\FOne}[1]{F_{#1}}

\newcommand{\DefineAttribute}[2]{%
  \expandafter\DeclareMathOperator\csname Op#1\endcsname{\mathit{#2}}
  \expandafter\newcommand\csname #1\endcsname[1]{\csname Op#1\endcsname(##1)}
}

\DefineAttribute{Frames}{frames}
\DefineAttribute{Framerate}{fps}
\DefineAttribute{Dataset}{dataset}
\DefineAttribute{Start}{start}
\DefineAttribute{End}{end}
\DefineAttribute{Width}{width}
\DefineAttribute{Height}{height}
\DefineAttribute{Channels}{channels}
\DefineAttribute{Distance}{distance}

\usepackage{acro}

\DeclareAcronym{LASSO}{
  short=LASSO,
  long=least absolute shrinkage and selection operator
}

\DeclareAcronym{AVA}{
  short=AVA,
  long=A Large-Scale Database for Aesthetic Visual Analysis,
  pdfstring=A Large-Scale Database for Aesthetic Visual Analysis (AVA)
}

\DeclareAcronym{CART}{
  short=CART,
  long=Classification And Regression Trees
}

\DeclareAcronym{HOG}{
  short=HOG,
  long=Classification And Regression Trees
}

\DeclareAcronym{FHOG}{
  short=FHOG,
  long=Felzenszwalb's HOG
}

\DeclareAcronym{LFW}{
  short=LFW,
  long=Labeled Faces in the Wild
}

\DeclareAcronym{SUMME}{
  short=SumMe,
  long=The SumMe Video Summarization
}

\DeclareAcronym{TVSUM50}{
  short=TVSum50,
  long=Summarizing Web Videos using Titles
}

\def\FRThreshold{\tau}
\def\FRThresholdValue{0.6}

\colorlet{DecisionGood}{green!80!black}
\colorlet{DecisionBad}{red!80!black}
\tikzstyle{Data}=[]
\tikzstyle{Process}=[draw, rectangle, align=center]
\tikzstyle{decision node}=[draw=black!30, rectangle, rounded corners]
\tikzstyle{decision label}=[fill=white, rectangle, font=\small]
\tikzstyle{decision end 0}=[draw=DecisionBad, font=\color{DecisionBad}, rectangle]
\tikzstyle{decision end 1}=[draw=DecisionGood, font=\color{DecisionGood}, rectangle]

\usepackage{algorithm}
\usepackage{algpseudocode}
\usepackage{booktabs}
\usepackage{multirow}
\usepackage{mathtools}
\DeclarePairedDelimiter{\Cardinality}{\lvert}{\rvert}%
\newcommand{\SetWhere}[2]{\left\{ #1\ \middle\vert\ #2 \right\}}
\newcommand{\Set}[1]{\{#1\}}

\DeclareMathOperator*{\argmin}{arg\,min}
\DeclareMathOperator*{\argmax}{arg\,max}
\DeclareMathOperator*{\mean}{mean}

\newcommand{\Sep}[2][\quad]{#1\text{#2}#1}
\definecolor{MyRed}{HTML}{E41A1C}
\definecolor{MyBlue}{HTML}{377EB8}
\definecolor{MyGreen}{HTML}{4DAF4A}
\definecolor{MyPurple}{HTML}{984EA3}
\definecolor{MyOrange}{HTML}{FF7F00}
\definecolor{MyYellow}{HTML}{FFFF33}
\definecolor{MyBrown}{HTML}{A65628}
\definecolor{MyPink}{HTML}{F781BF}
\definecolor{MyGray}{HTML}{999999}
\colorlet{plota}{MyRed}
\colorlet{plotb}{MyBlue}
\colorlet{plotc}{MyGreen}
\colorlet{plotd}{MyPurple}
\colorlet{plote}{MyOrange}
\colorlet{plotf}{MyYellow}
\colorlet{plotg}{MyBrown}
\colorlet{ploth}{MyPink}
\colorlet{ploti}{MyGray}
\newcommand{\Matrix}[1]{\mathbf{#1}}
\newcommand{\SetV}[1]{\mathbb{#1}}
\DeclarePairedDelimiter{\norm}{\lVert}{\rVert}
\usepackage{amsfonts}
\newcolumntype{P}[1]{>{\raggedright\arraybackslash}p{#1}}

\DeclareMathOperator{\Mean}{mean}
\DeclareMathOperator{\Stddev}{std}

\usepackage{makecell}
\usepackage{colortbl}
\usepackage{textcomp}
\def\FrameLabelingBlurryLuminance{0.45}
\def\FrameLabelingBlurrySharpness{83.19}
\def\FrameLabelingBlurryUniformity{0.75}
\def\FrameLabelingDarkLuminance{0.02}
\def\FrameLabelingDarkSharpness{3.42}
\def\FrameLabelingDarkUniformity{0.15}
\def\FrameLabelingGoodLuminance{0.63}
\def\FrameLabelingGoodSharpness{1955.33}
\def\FrameLabelingGoodUniformity{0.70}
\def\FrameLabelingUniformLuminance{0.14}
\def\FrameLabelingUniformSharpness{6552.91}
\def\FrameLabelingUniformUniformity{0.24}
\newcommand{\FMeasureTop}[1]{\cellcolor[HTML]{31A354}\textbf{#1}}
\newcommand{\FMeasureSecond}[1]{\cellcolor[HTML]{ADDD8E}\textbf{#1}}
\newcommand{\FMeasureThird}[1]{\cellcolor[HTML]{D2FFB5}\textbf{#1}}
\def\FMeasureTable{
\begin{tabular}{lcccccccccc}
\toprule
&\multicolumn{2}{c}{\textbf{Dataset}} & \multicolumn{3}{c}{\textbf{Humans}} & \multicolumn{5}{c}{\textbf{Computational Methods}} \\ \cmidrule(r){2-3} \cmidrule(r){4-6} \cmidrule(r){7-11}
\textbf{Videoname} & \textbf{Random} & \textbf{Upper Bound} & \textbf{Worst} & \textbf{Mean} & \textbf{Best} & \textbf{Uniform} & \textbf{Cluster} & \textbf{Attn.} & \textbf{Summe} & \textbf{Ours}\\
\midrule
Air Force One & 0.144 & 0.490 & 0.185 & 0.332 & 0.457 & 0.161 & 0.143 & \FMeasureThird{0.215} & \FMeasureSecond{0.318} & \FMeasureTop{0.362}\\
Base jumping & 0.144 & 0.398 & 0.113 & 0.257 & 0.396 & \FMeasureSecond{0.168} & 0.109 & \FMeasureTop{0.194} & \FMeasureThird{0.121} & 0.106\\
Bearpark climbing & 0.147 & 0.330 & 0.129 & 0.208 & 0.267 & 0.152 & \FMeasureThird{0.158} & \FMeasureSecond{0.227} & 0.118 & \FMeasureTop{0.261}\\
Bike Polo & 0.134 & 0.503 & 0.190 & 0.322 & 0.436 & 0.058 & \FMeasureThird{0.130} & 0.076 & \FMeasureTop{0.356} & \FMeasureSecond{0.301}\\
Bus in Rock Tunnel & 0.135 & 0.359 & 0.126 & 0.198 & 0.270 & \FMeasureThird{0.124} & 0.102 & 0.112 & \FMeasureSecond{0.135} & \FMeasureTop{0.147}\\
Car railcrossing & 0.140 & 0.515 & 0.245 & 0.357 & 0.454 & \FMeasureThird{0.146} & 0.146 & 0.064 & \FMeasureTop{0.362} & \FMeasureSecond{0.192}\\
Cockpit Landing & 0.136 & 0.443 & 0.110 & 0.279 & 0.366 & 0.129 & \FMeasureThird{0.156} & 0.116 & \FMeasureSecond{0.172} & \FMeasureTop{0.201}\\
Cooking & 0.145 & 0.528 & 0.273 & 0.379 & 0.496 & \FMeasureThird{0.171} & 0.139 & 0.118 & \FMeasureSecond{0.321} & \FMeasureTop{0.348}\\
Eiffel Tower & 0.130 & 0.467 & 0.233 & 0.312 & 0.426 & \FMeasureThird{0.166} & \FMeasureSecond{0.179} & 0.136 & \FMeasureTop{0.295} & 0.088\\
Excavators river crossing & 0.144 & 0.411 & 0.108 & 0.303 & 0.397 & 0.131 & \FMeasureThird{0.163} & 0.041 & \FMeasureSecond{0.189} & \FMeasureTop{0.231}\\
Fire Domino & 0.145 & 0.514 & 0.170 & 0.394 & 0.517 & \FMeasureThird{0.233} & \FMeasureTop{0.349} & \FMeasureSecond{0.252} & 0.130 & 0.169\\
Jumps & 0.149 & 0.611 & 0.214 & 0.483 & 0.569 & 0.052 & \FMeasureThird{0.298} & 0.243 & \FMeasureSecond{0.427} & \FMeasureTop{0.542}\\
Kids playing in leaves & 0.139 & 0.394 & 0.141 & 0.289 & 0.416 & \FMeasureTop{0.209} & \FMeasureSecond{0.165} & 0.084 & 0.089 & \FMeasureThird{0.093}\\
Notre Dame & 0.137 & 0.360 & 0.179 & 0.231 & 0.287 & 0.124 & \FMeasureSecond{0.141} & \FMeasureThird{0.138} & \FMeasureTop{0.235} & 0.107\\
Paintball & 0.127 & 0.550 & 0.145 & 0.399 & 0.503 & 0.109 & 0.198 & \FMeasureSecond{0.281} & \FMeasureTop{0.320} & \FMeasureThird{0.213}\\
Playing on water slide & 0.134 & 0.340 & 0.139 & 0.195 & 0.284 & \FMeasureThird{0.186} & 0.141 & 0.124 & \FMeasureSecond{0.200} & \FMeasureTop{0.218}\\
Saving dolphines & 0.144 & 0.313 & 0.095 & 0.188 & 0.242 & \FMeasureSecond{0.165} & \FMeasureTop{0.214} & \FMeasureThird{0.154} & 0.145 & 0.128\\
Scuba & 0.138 & 0.387 & 0.109 & 0.217 & 0.302 & \FMeasureThird{0.162} & 0.135 & \FMeasureTop{0.200} & \FMeasureSecond{0.184} & 0.140\\
St Maarten Landing & 0.143 & 0.624 & 0.365 & 0.496 & 0.606 & 0.092 & 0.096 & \FMeasureSecond{0.419} & \FMeasureThird{0.313} & \FMeasureTop{0.557}\\
Statue of Liberty & 0.122 & 0.332 & 0.096 & 0.184 & 0.280 & \FMeasureThird{0.143} & 0.125 & 0.083 & \FMeasureSecond{0.192} & \FMeasureTop{0.259}\\
Uncut Evening Flight & 0.131 & 0.506 & 0.206 & 0.350 & 0.421 & \FMeasureThird{0.122} & 0.098 & \FMeasureTop{0.299} & \FMeasureSecond{0.271} & 0.081\\
Valparaiso Downhill & 0.142 & 0.427 & 0.148 & 0.272 & 0.400 & 0.154 & 0.154 & \FMeasureThird{0.231} & \FMeasureSecond{0.242} & \FMeasureTop{0.288}\\
car over camera & 0.134 & 0.490 & 0.214 & 0.346 & 0.418 & 0.099 & \FMeasureThird{0.296} & 0.201 & \FMeasureSecond{0.372} & \FMeasureTop{0.408}\\
paluma jump & 0.139 & 0.662 & 0.346 & 0.509 & 0.642 & \FMeasureThird{0.132} & 0.072 & 0.028 & \FMeasureSecond{0.181} & \FMeasureTop{0.334}\\
playing ball & 0.145 & 0.403 & 0.190 & 0.271 & 0.364 & \FMeasureTop{0.179} & \FMeasureSecond{0.176} & 0.140 & \FMeasureThird{0.174} & 0.151\\
\midrule
Average & 0.139 & 0.454 & 0.179 & 0.311 & 0.409 & 0.143 & 0.163 & \FMeasureThird{0.167} & \FMeasureSecond{0.234} & \FMeasureTop{0.237}\\
\bottomrule
\end{tabular}
}
\def\PerformanceTable{
\begin{tabular}{lrrrrrrr}
\toprule
\textbf{Video Name} & \textbf{Duration (s)} & \textbf{Time (s)} & \textbf{Speed}\\
\midrule
Jumps & 38.00 & 19.12 & 1.99x \\
Cooking & 85.80 & 22.16 & 3.87x \\
Fire Domino & 53.73 & 27.99 & 1.92x \\
St Maarten Landing & 70.04 & 36.72 & 1.91x \\
Scuba & 74.03 & 48.45 & 1.53x \\
paluma jump & 85.89 & 46.89 & 1.83x \\
Bike Polo & 102.13 & 69.50 & 1.47x \\
Playing on water slide & 102.27 & 54.76 & 1.87x \\
playing ball & 103.97 & 54.52 & 1.91x \\
Kids playing in leaves & 106.34 & 71.29 & 1.49x \\
Bearpark climbing & 133.64 & 78.31 & 1.71x \\
Statue of Liberty & 154.52 & 69.89 & 2.21x \\
car over camera & 146.21 & 71.04 & 2.06x \\
Air Force One & 179.76 & 103.59 & 1.74x \\
Notre Dame & 192.00 & 106.87 & 1.80x \\
Base jumping & 157.79 & 105.27 & 1.50x \\
Eiffel Tower & 198.84 & 118.90 & 1.67x \\
Car railcrossing & 169.34 & 115.14 & 1.47x \\
Bus in Rock Tunnel & 171.10 & 109.00 & 1.57x \\
Valparaiso Downhill & 172.77 & 115.51 & 1.50x \\
Paintball & 254.25 & 137.37 & 1.85x \\
Saving dolphines & 222.99 & 120.15 & 1.86x \\
Cockpit Landing & 301.83 & 200.50 & 1.51x \\
Uncut Evening Flight & 322.72 & 215.42 & 1.50x \\
Excavators river crossing & 388.84 & 210.87 & 1.84x \\
\midrule
Average &&& 1.82x \\
\bottomrule
\end{tabular}
}

\usepackage[capitalise,noabbrev]{cleveref}

\newcommand{\PlotOurs}[2]{
\begin{tikzpicture}
\begin{axis}[scale only axis, width=0.98\textwidth, height=0.1\textheight, table/col sep=comma, ymin=-0.1]
\addplot+ [color=MyBlue, smooth, no marks, ultra thick] table [x=frame, y=value] {data/#1-ours.csv};
 \foreach \ss/\ee in {#2} {
   \edef\temp{\noexpand\fill [MyGreen, opacity=0.4] (axis cs:\ss, -0.1) rectangle (axis cs:\ee, 2);}
   \temp
}
\end{axis}
\node [inner sep=1pt, anchor=north west, font=\sf\scriptsize\color{gray}] at (current bounding box.north west) {\textbf{Our work}};
\end{tikzpicture}
}

\setcopyright{rightsretained}

\acmDOI{10.475/123_4}

\acmISBN{123-4567-24-567/08/06}

\acmConference[ICDSC'18]{ACM International Conference on Distributed Smart Cameras}{September 2018}{Eindhoven, Netherlands}
\acmYear{2018}
\copyrightyear{2018}

\acmArticle{4}
\acmPrice{15.00}

\begin{document}

\copyrightyear{2018} 
\acmYear{2018} 
\setcopyright{acmcopyright}
\acmConference[ICDSC '18]{International Conference on Distributed Smart Cameras}{September 3--4, 2018}{Eindhoven, Netherlands}
\acmBooktitle{International Conference on Distributed Smart Cameras (ICDSC '18), September 3--4, 2018, Eindhoven, Netherlands}
\acmPrice{15.00}
\acmDOI{10.1145/3243394.3243689}
\acmISBN{978-1-4503-6511-6/18/09}

\title{Real-time Video Summarization on Commodity Hardware}

\author{Wesley Taylor and Faisal Z. Qureshi}
\affiliation{%
  \institution{Faculty of Science, University of Ontario Institute of Technology}
  \streetaddress{UA4000, 2000 Simcoe St. N.}
  \city{Oshawa}
  \state{ON L1G 0C5 Canada}
}
\email{{wesley.taylor3|faisal.qureshi}@uoit.net}

\renewcommand{\shortauthors}{Taylor and Qureshi}

\begin{abstract}
  We present a method for creating video summaries in real-time on commodity hardware.  Real-time here refers to the fact that the time required for video summarization is less than the duration of the input video.  First, low-level features are use to discard undesirable frames.  Next, video is divided into segments, and segment-level features are extracted for each segment.  Tree-based models trained on widely available video summarization and computational aesthetics datasets are then used to rank individual segments, and top-ranked segments are selected to generate the final video summary.  We evaluate the proposed method on \ac{SUMME} dataset and show that our method is able to achieve summarization accuracy that is comparable to that of a current state-of-the-art deep learning method, while posting significantly faster run-times.  Our method on average is able to generate a video summary in time that is shorter than the duration of the video.
\end{abstract}

%
%



\maketitle

\section{Introduction}

Cameras are now ubiquitous.  This has resulted in an explosive growth
in user-generated images and videos.  In the case of videos, at least,
our ability to record videos has far outpaced methods and tools to
manage these videos.  A skier, for example, can easily record many
hours of video footage using an action camera, such as a GoPro.  Raw
video footage, in general, is {\it unviewable}---the recorded video
needs to be summarized or edited in some manner before it can be
shared with others. Clearly, no one is interested in watching many
hours of skiing video when most of it is bound to be highly
repetitive.  Manual video editing and summarizing is painstakingly
slow and tedious.  Consequently a large fraction of recorded footage
is never shared or even viewed.  We desperately need one-touch video
editing tools capable of generating video summarizes that capture the
meaningful and interesting portions of the video, discarding sections
that are boring, repetitive or poorly recorded.  Such tools will
revolutionize how we share video stories with friends and family via social
media.

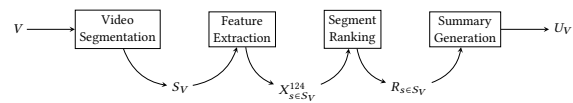
\begin{figure}[b]
\centering
\small
\resizebox{0.9\columnwidth}{!}{
\begin{tikzpicture}[->,>=stealth]
\node[Data] (video) {$\Video$};
\node[Process, right=of video] (segmentation process) {Video\\Segmentation};
\node[Process, right=of segmentation process] (feature extraction process) {Feature\\Extraction};
\node[Process, right=of feature extraction process] (segment ranking process) {Segment\\Ranking};
\node[Process, right=of segment ranking process] (summary generation process) {Summary\\Generation};
\coordinate (temp) at ($(segmentation process)!0.5!(feature extraction process)$);
\node[Data, below=of temp] (segmentation) {$\Segmentation[\Video]$};
\coordinate (temp) at ($(feature extraction process)!0.5!(segment ranking process)$);
\node[Data, below=of temp] (features) {$\Features{124}{\Segment\In\Segmentation[\Video]}$};
\coordinate (temp) at ($(segment ranking process)!0.5!(summary generation process)$);
\node[Data, below=of temp] (ranks) {$\Ranking[{\Segment\In\Segmentation[\Video]}]$};
\node[Data, right=of summary generation process] (summary) {$\Summary[\Video]$};
\draw (video) to (segmentation process);
\draw (segmentation process) to [bend right] (segmentation);
\draw (segmentation) to [bend right] (feature extraction process);
\draw (feature extraction process) to [bend right] (features);
\draw (features) to [bend right] (segment ranking process);
\draw (segment ranking process) to [bend right] (ranks);
\draw (ranks) to [bend right] (summary generation process);
\draw (summary generation process) to (summary);
\end{tikzpicture}
}
\caption{Our summarization pipeline.  Video $\Video$ is processed to
  construct a set of segments $\Segmentation[\Video]$.  Next, a
  124-dimensional feature vector $\Features{124}{s}$ is extracted for
  each segment $\Segment \In \Segmentation[\Video]$.  These features
  are processed to assign a ranking to each segment $\Segment \In
  \Segmentation[\Video]$.  Final step consists of selecting the top
  ranked segments to create the summary $\Summary[\Video]$.}
\label{fig:system-overview}
\end{figure}

A meaningful video summarization needs to take into account both the
user context and the video content.  Two different users may find
entirely different sections of a recorded video interesting.
Consider, for example, the scenario where someone records a children
soccer match.  Parents may only be interested in a section in video
that shows their child.  We refer to this as user context.  Video
summarization algorithms, therefore, should take into account the
likes and dislikes of the viewers of the video summary.  Video content
is also important.  By necessity video summarization algorithms relies
upon video content to select which portions of the videos {\it make
  the cut}.

This paper develops a real-time video summarization system (Figure~\ref{fig:system-overview}).  The
proposed system is able to perform video summarization at speeds that
far exceed those achieved by state-of-the-art deep learning approaches
for video summarization.  We list these approaches in the next
section.  The proposed system exploits low-level image features to
efficiently discard segments with {\it low interestingness} or having
{\it poor quality}.  This means that subsequent summarization steps,
which are computationally expensive, only deal with the remaining
segments.  This can lead to significant savings, especially for long
duration videos, such as the all day ski trip video in the example
mentioned above.  A key feature of the proposed system is its ability
to generate alternate summaries almost instantaneously.  A user can
guide the system to generate a different summary thereby injecting
user-preference into the process of summarization.
\cref{fig:system-overview} shows our summarization pipeline.

We evaluate the proposed method on SumMe video summarization
benchmark, and compare our method with a number of existing video
summarization schemes.  Our method achieves the highest $F_1$-measure.
It also achieves highest accuracy on over 50\% of the tested videos.
We also show that the summarization times of the proposed method
increases linearly with the duration of the input video.

The rest of the paper is organized as follows.  We briefly discuss
related work in the next section.  \cref{sec:video-segmentation}
discusses video segmentation.  Segment ranking is covered in
\cref{sec:feature-extraction}.  The following section describes
video summarization.  We conclude the paper with evaluation and
results and conclusions in the last two sections.

\newcommand{\DrawFrame}[3][]{%
  \draw[black] ($(base)+(#2-1, 0)$) rectangle ($(base)+(#2, 1)$);
  \node[anchor=center] at ($(base)+(#2-1,0)+(0.5,0.5)$) {#3};
  \node[anchor=north, font=\scriptsize] at ($(base)+(#2-1,0)+(0.5,0)$) {#1};
}
\newcommand{\BetweenStepArrow}[1][8.5]{\node[shape border rotate=270, draw, single arrow, anchor=tail, minimum height=1.6cm, text width=1cm, inner sep=0cm] at (#1,-0.5) {};}
\newcommand{\BetweenStepArrowShifted}[2]{\node[shape border rotate=270, draw, single arrow, anchor=tail, minimum height=1.6cm, text width=1cm, inner sep=0cm] at (#1,#2) {};}
\newcommand{\HighlightFrame}[1]{%
  \draw[black, fill=black!20!white] ($(current)+(#1,0)$) rectangle ($(current)+(#1+1,1)$);}
\newcommand{\DrawCurrentFrame}{\draw[ultra thick, black] ($(current)$) rectangle ($(current)+(1,1)$);}

\begin{figure}[b]
\centering
\begin{minipage}[t]{0.23\columnwidth}
\centering
\includegraphics[width=\textwidth]{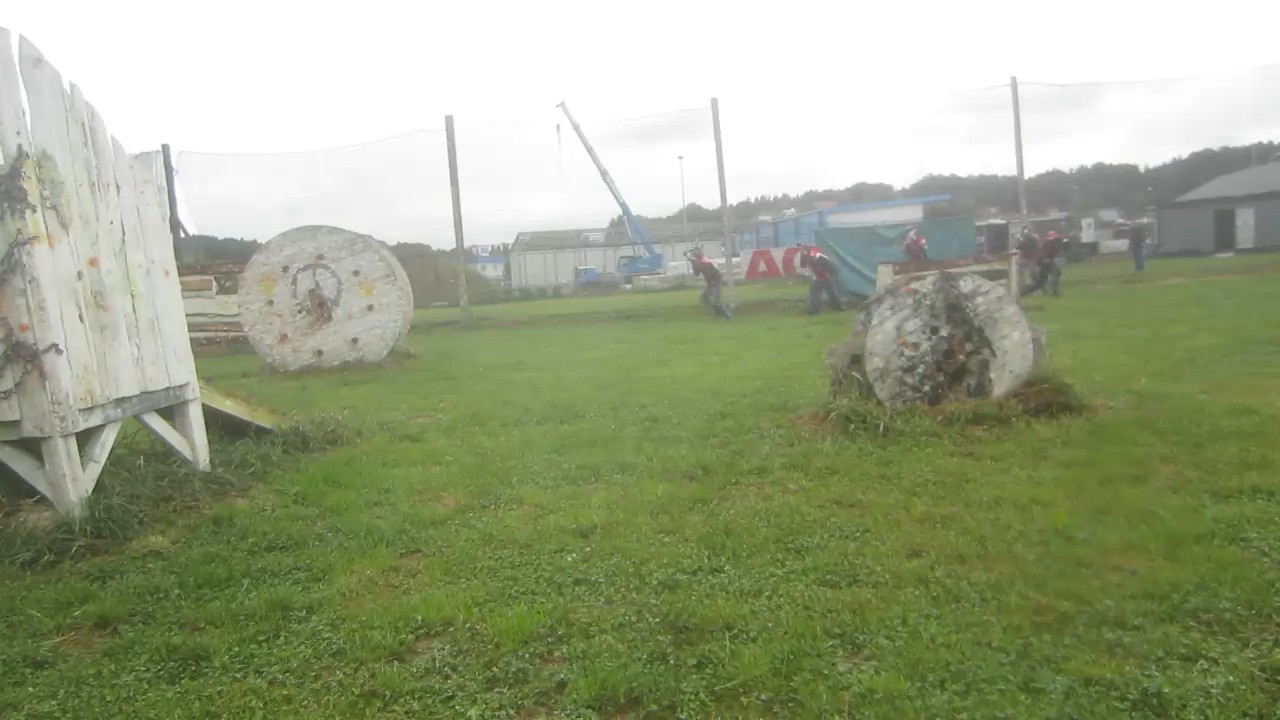} \\[1ex]
\begin{tabular}{l@{ $=$ }r}
  $\Luminance$  & $\FrameLabelingGoodLuminance$ \\
  $\Sharpness$  & $\FrameLabelingGoodSharpness$ \\
  $\Uniformity$ & $\FrameLabelingGoodUniformity$
\end{tabular}\\
Label: \emph{none}
\end{minipage}
\hfill
\begin{minipage}[t]{0.23\columnwidth}
\centering
\includegraphics[width=\textwidth]{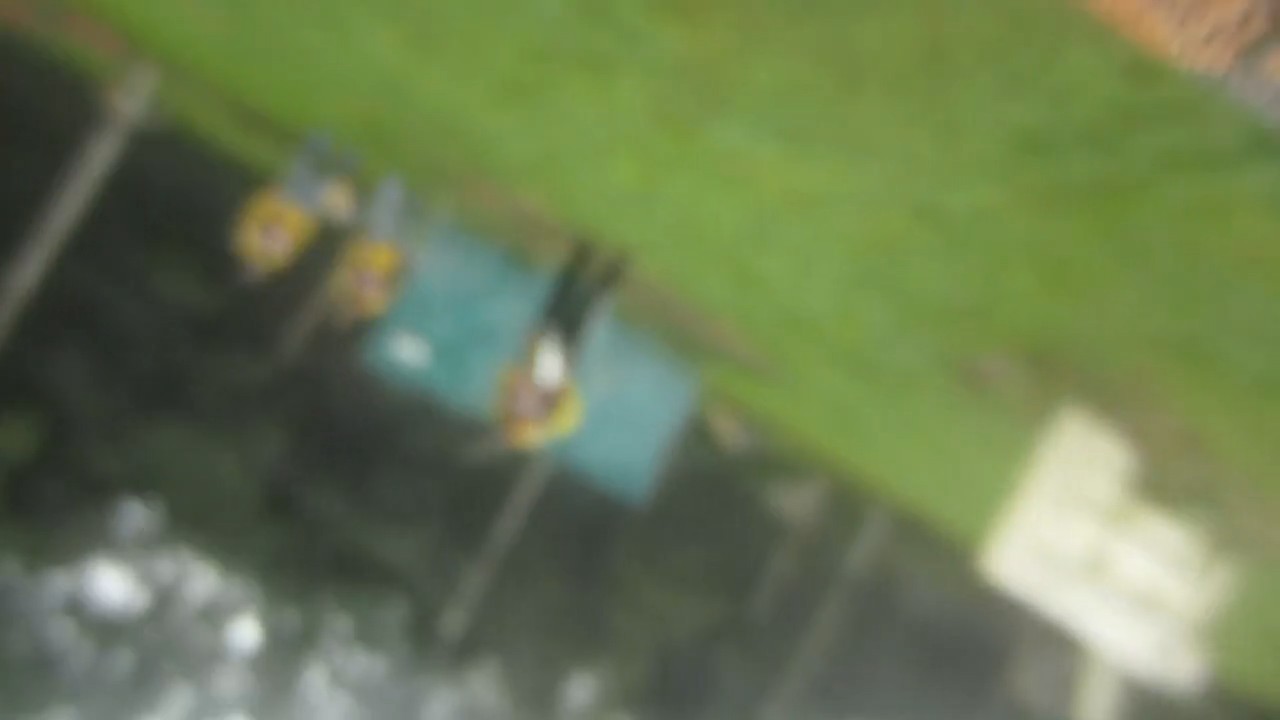}\\[1ex]
\begin{tabular}{l@{ $=$ }r}
  $\Luminance$  & $\FrameLabelingBlurryLuminance$ \\
  $\Sharpness$  & \color{red} $\FrameLabelingBlurrySharpness$ \\
  $\Uniformity$ & $\FrameLabelingBlurryUniformity$
\end{tabular} \\
Label: \textbf{blurry}
\end{minipage}
\hfill
\begin{minipage}[t]{0.23\columnwidth}
\centering
\includegraphics[width=\textwidth]{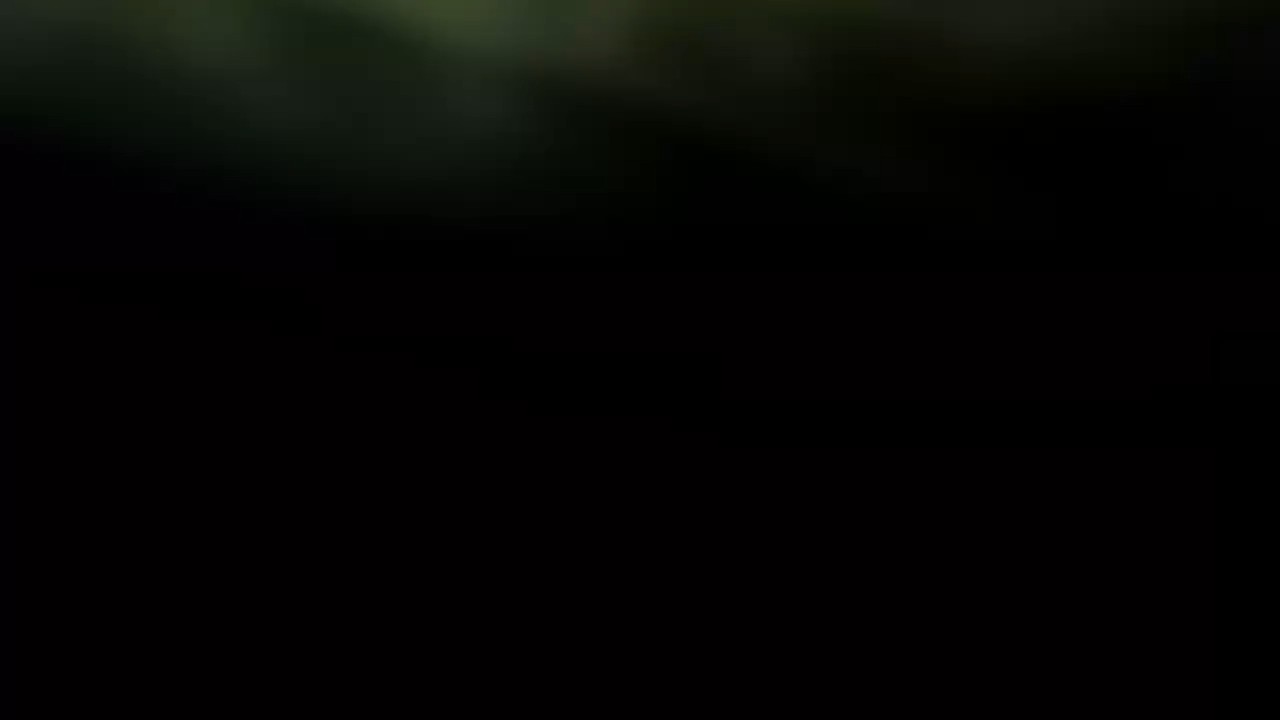}\\[1ex]
\begin{tabular}{l@{ $=$ }r}
  $\Luminance$  & \color{red} $\FrameLabelingDarkLuminance$ \\
  $\Sharpness$  & \color{red} $\FrameLabelingDarkSharpness$ \\
  $\Uniformity$ & \color{red} $\FrameLabelingDarkUniformity$
\end{tabular} \\
Label: \textbf{dark}
\end{minipage}
\hfill
\begin{minipage}[t]{0.23\columnwidth}
\centering
\includegraphics[width=\textwidth]{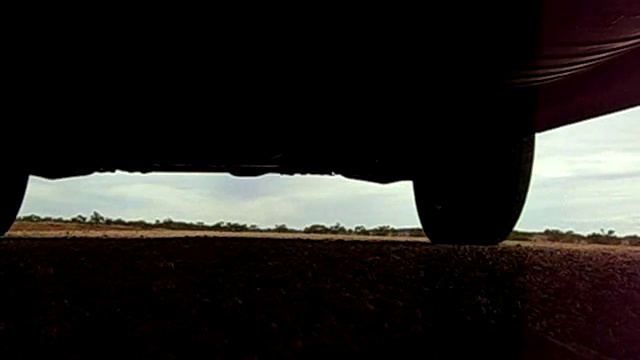}\\[1ex]
\begin{tabular}{l@{ $=$ }r}
  $\Luminance$  & $\FrameLabelingUniformLuminance$ \\
  $\Sharpness$  & $\FrameLabelingUniformSharpness$ \\
  $\Uniformity$ & \color{red} $\FrameLabelingUniformUniformity$
\end{tabular} \\
Label: \textbf{uniform}
\end{minipage}
\caption{Using luminance ($\FrameLabelingDarkLuminance$), sharpness ($\FrameLabelingDarkSharpness$) and uniformity
  ($\FrameLabelingDarkUniformity$) to label undesirable frames.  The
  values shown in red indicates that these fall below the empirically
  selected threshold values.}
\label{fig:frame-labels}
\end{figure}

\section{Background}\label{sec:background}

A majority of the existing video summarization methods follow a common
recipe: step 1) video segmentation, step 2) segment ranking and step
3) segment selection~\cite{clustering,attention,SUMME,Yahoo}.  Methods
vary in how segmentation is performed and how individual segments are
ranked.  \cite{LSTM} is an exception to this rule that uses recent
advances in deep learning and provides an end-to-end system for video
summarization.  This method relies upon the availability of suitable
training data.  Early video summarization methods were unsupervised
\cite{clustering, attention, Yahoo}; however, with the recent
availability of high-quality video summarization datasets, many newer
methods are supervised~\cite{SUMME,LSTM}.

Video summarization has also been explored in the context of
robotics~\cite{GirPhD14}.  Their motivation stems from the
fact that transmitting raw video footage, say to a base station,
incurs large communication costs.  It is also infeasible in
situations where bandwidth is limited.  They leverage topic modeling
to identify the novel segments of the recorded video with a view to
construct a video summarization that captures the salient pieces of
the video. 

Clustering~\cite{clustering} and attention~\cite{attention} methods
are often used as baselines when evaluating new summarization methods.
The first method performs clustering to get segmentation, and uses a
0/1 knapsack for segment selection for final summary generation.  The
second method extracts attention features for each frame, assigning an
interestingness score to each frame.  Frames with high interestingness
scores are selected to generate the summary.  We refer the kind reader
to the respective publications for technical details.  Suffice to say
that both classes of methods are unsupervised and are able to achieve higher
accuracy when compared to a method that picks frames (or segments) at
random when generating a video summarization.  Recent methods outperform
both these methods.

Method developed in~\cite{SUMME} is of particular interest to us.
\cite{SUMME} not only developed a new method for video summarization.
It also created a first-of-its-kind benchmark for video summarization.
This dataset is referred to as the SumMe dataset.  We too use this
dataset to evaluate the performance of our method.  \cite{SUMME} uses
change point detection for segmentation.  These segments are
subsequently ranked and the final summary is generated using a 0/1
knapsack formulation.  \cite{Yahoo} method is similar to the method
proposed in \cite{SUMME}.  The key difference is that~\cite{Yahoo}
method uses a different set of features for ranking segments.

The current best performing video summarization method is~\cite{LSTM}.
It uses convolutional and recurrant layers that operate upon sequences
of frames and compute interestingness score for each frame.
Specifically, this method uses pool-5 layer of GoogLeNet model as
frame-level features, which are fed into LSTM units to generate frame
and segment level interestingness scores.  The key idea is to capture
temporal relationship between successive frames to compute frame-level
interestingness score suitable for video summarization.

\section{Video Segmentation}\label{sec:video-segmentation}

The algorithm begins by identifying frames that are too dark, blurry,
or uniform (see \cref{fig:frame-labels}).  Luminance ($\Luminance$), sharpness ($\Sharpness$) and
uniformity ($\Uniformity$) values are computed for each frame to label the frame accordingly.
Luminance is given by
$$\Luminance = \mean(0.2126\cdot\Red + 0.7152\cdot\Green + 0.0722\cdot\Blue),$$
sharpness is computed as
$$\Sharpness = \mean(\Gradient{x}^2 + \Gradient{y}^2), \mathrm{and}$$
uniformity value is computed by first constucting a normalized 1D
grayscale histgoram $\Histogram$ with 128 bins and then computing the ratio between
the top $5^{\text{th}}$ percentile bins of $\Histogram$ and the rest
of $\Histogram$.  These features have low computational overhead.  The
algorithm thus avoids wasting precious computational resources (during
the subsequent steps) on frames that will not make the final cut any
ways.

Next, input video $\Video$ is divided into one or more non-overlapping segments
$$\Segmentation[\Video] = \Set{\Segment[0],\ldots,\Segment[k]}.$$
While these segments do not overlap, we allow for gaps between
adjacent segments, i.e.,  we only require that $\End{\Segment[i]} < \Start{\Segment[i+1]}$.  We formulate our video as a 
multidimensional time-series, allowing us to cast video segmentation
as a multiple change point detection problem \cite{changepointlasso,TVSUM}.

Change point detection operates upon a time series feature matrix
$\Matrix{X}$, where column $i$ stores features extracted from frame $i$.
Our method extracts $2200$-dimensional
feature vector from each video frame.  Specifically each frame is
represented using a HSV histogram with 128 bins per channel and an
edge orientation and magnitude histogram with 30 bins each.  These
features are extracted over a two-level pyramid consisting of 5
regions, which yields a $2200$-dimensional feature.  Each video is now
represented as a $2200 \times n$ matrix $\Matrix{X}$.  Here $n$ indicates the
number of frames. A set of sparse
coefficients $\Matrix{A} \In \SetV{R}^{n\times{}n}$ is computed from
$\Matrix{X}$ by solving the following convex optimization problem.
\begin{equation} \label{eq:lasso}
\argmin_{\Matrix{A}} \norm{\Matrix{X}-\Matrix{XA}}_\text{F}^2 + \frac{\lambda}{2} \norm{\Matrix{A}}_{2,1}.
\end{equation}
$\Matrix{A}$ is used to assign a score to each frame, and the top
ranked $k$ frames are selected as split points to generate $k+1$
segments.  We set the problem so that average segment duration is roughly
$5$ seconds.
This method can be thought of as a more robust version of
threshold-based and content-aware sampling. Rather than relying simply
on local color or brightness features, a combination of color and edge
histograms are used to locate segment boundaries based on the
statistical properties of the entire video.

Next we refine the segmentation by removing dark, blurry or uniform
frames.  Segments having a large fraction of undesirable frames are
discarded in the process, which also results in further savings down
the line.  Segments can also be trimmed, discarding undesirable frames
at the either end, or split into two or more segments. Adjacent
segments containing too few frames are also merged to form a single
segment at this stage.  This process is shown in Algorithm~\ref{alg:segment-merging-code}.

\begin{algorithm}[t]
\linespread{1.2}\selectfont
\vspace{1ex}
\begin{algorithmic}[1]
\Statex \textbf{Inputs:}
\Statex \parbox{0.05\textwidth}{\hfill{}$\Segmentation$}: A segmentation consisting of $n$ segments $\{\Segment[0],\ldots,\Segment[n-1]\}$
\Statex \parbox{0.05\textwidth}{\hfill{}$\Duration{m}$}: The minimum segment frame duration threshold
\Statex \parbox{0.05\textwidth}{\hfill{}$\Duration{b}$}: The between segment frame duration threshold
\Function{PostProcessShortSegments}{$\Segmentation$, $\Duration{m}$, $\Duration{b}$}
\For {$\Segment[p],\Segment,\Segment[n]$ in $\Call{Zip}{\Segmentation,\Segmentation{}[1\,:],\Segmentation{}[2\,:]}$}
  \If {$\Frames{s} > \Duration{m}$}
    \State \textbf{continue}
  \EndIf
  \State $\text{merged} \gets \Call{False}{}$
  \If {$\Distance{\Segment[p],\Segment} \leq \Duration{b}$}
    \State $\Segmentation \gets \Call{Remove}{\Segmentation,\Segment[p]}$
    \State $\Start{\Segment} \gets \Start{\Segment[p]}$
    \State $\text{merged} \gets \Call{True}{}$
  \EndIf
  \If {$\Distance{\Segment,\Segment[n]} \leq \Duration{b}$}
    \State $\Segmentation \gets \Call{Remove}{\Segmentation,\Segment[n]}$
    \State $\End{\Segment} \gets \End{\Segment[n]}$
    \State $\text{merged} \gets \Call{True}{}$
  \EndIf
  \If {$\text{merged} = \Call{False}{}$}
    \State $\Segmentation \gets \Call{Remove}{\Segmentation,\Segment}$
  \EndIf
\EndFor
\State \Return $\Segmentation$
\EndFunction
\Statex \textbf{Output:} A new version of $\Segmentation$ with segment merging and elimination applied
\end{algorithmic}
\caption[Segment Merging Algorithm]{The algorithm used for performing segment merging and elimination.}
\label{alg:segment-merging-code}
\end{algorithm}

\section{Segment Ranking}\label{sec:feature-extraction}

Once all candidate segments $\Segmentation[\Video]$ for our video
$\Video$ have been located, the next step is to rank these segments.
The algorithm begins by extracting frame-wise features, which are subsequently
used to rank the individual segments.

\subsection{Frame-Level Features}

We compute a 62 dimensional
feature vector $\Features{62}{f}$ for each frame as follows.
The first 59 dimensions correspond to computational aesthetic features
computed at each frame (Table~\ref{tab:computational-aesthetics}).  We refer the interested reader to
\cite{aesthetics1,aesthetics2,aesthetics3,aesthetics4} for technical
details about these features.
Dimensions 60 contains the number of faces seen in this frame, and
dimension 61 records the number of ``salient'' faces seen in this
frame.  The last dimension stores a 1 if the frame is deemed
aesthetically pleasing (see below).

\subsubsection{Salient Face Detection}

The process of finding salient faces consists of three steps:
a) face detection, b) (face) feature vector extraction, and c) (face)
clustering.  The algorithm employs \ac{FHOG} for face
detection~\cite{facedetect}.  To extract a face feature vector, we
employ a modfied version of ResNet-34~\cite{resnet}, containing
only 29 layers and half the number of filters in each layer.  We train
the network using a metric loss function over 3 million faces from the
FaceScrub~\cite{facescrub} and VGG-Face~\cite{vggface} datasets.  This
model is able to predict with $99.38\%$ accuracy if two faces belong
to the same individual on the \ac{LFW}~\cite{lfw} dataset.

Face
feature vectors are clustered using Chinese whispers graph clustering
algorithm~\cite{chinesewhispers}.  Chinese whispers is a linear-time
hard partitioning, randomized, flat clustering method.  A linear-time
algorithm is highly desirable since an hour long video can easily
contain more than 50,000 face feature vectors.  Clustering ensures
that each ``person'' ends up in at most one cluster.  Clusters with
large memberships identify salient persons.  Note that this method
requires no prior knowledge about salient faces.

The initial
``graph'' used as input to the clustering algorithm is constructed by
simply looping over every pair of features
$$\SetWhere{\left(\Features{128}{\Frame[a]},
  \Features{128}{\Frame[b]}\right)}{\Frame[a],\Frame[b]\In\Frames{\Segment},\Frame[a]\neq\Frame[b]}$$
\noindent across all segments and frames computed in the previous step, and
creating an ``edge'' between two nodes when their distance is below
some threshold value $\FRThreshold$. A value of $\FRThreshold =
\FRThresholdValue$ was selected, as it matches the value that was used
for the metric loss layer of the deep neural network used in the
previous step.

\subsubsection{Aesthetic Score}

The last dimension contains an aesthetic score of 0 or 1 for this
frame.  We use an XGBoost classifier trained on \ac{AVA} \cite{AVA}
dataset to compute this score.
Each image in \ac{AVA} dataset has an associated user score between 0
and 1, which captures the aesthetic appeal of that image.  For our
purposes, we assign a score of 0 for any image with ranking less than
0.5.  Images with ranking more than 0.5 are assigned a score of 1.  We
train an XBGoost classifier using 10-fold cross-validation and a
train/test split of $70\%/30\%$.  The input to this classifier are
computational aesthetic features listed in Table~\ref{tab:computational-aesthetics}.  The XGBoost
classifier obtains an accuracy of $73.66\%$, which is significantly
higher than the reference model shown in~\cite{AVA}.  The accuracy of
reference model is $53.85\%$.  ILGnet \cite{aestheticsmodel} posts
the current best accuracy of $82.66\%$.  ILGnet is a deep learning based
model, which is more tricky to train and has significantly worse
runtime performance than our XGBoost classifier.

\begin{table*}[t]
\centering
\renewcommand\arraystretch{1.4}
\begin{tabular}{lll}
  \toprule
  \textbf{Feature} & \textbf{Dim.} & \textbf{Description} \\
  \midrule
  Contrast & 1 & The ratio between the luminance range and average luminance. \\
  Image Mean HSV & 3 & The average H, S, and V values over the entire image. \\
  Center Mean HSV & 3 & The average H, S, and V values for the image center quadrant. \\
  Itten Histograms~\cite{aesthetics3} & 20 & Histograms of H values over 12 bins, S values over 5 bins, and V values over 3 bins. \\
  Itten Contrasts~\cite{aesthetics3}  & 3 & Standard deviation of each Itten Histogram. \\
  Pleasure, Arousal, Dominance~\cite{aesthetics3} & 3 & Approximate emotional values computed as linear combinations of the mean V and S values. \\
  Haralick Texture Features~\cite{haralick} & 13 & Average Haralick texture features over all four directions. \\
  Contrast Balance & 1 & Distance between the original and contrast-normalized grayscale image. \\
  Exposure Quality & 1 & Negative absolute value of luminance histogram skew. \\
  JPEG Quality~\cite{jpegquality} & 1 & No-reference quality estimation algorithm for JPEG images. \\
  Tenengrad~\cite{tenengrad} & 1 & Sharpness according to the Tenengrad method. \\
  Spectral Residual~\cite{spectralresidual} & 9 & Rule of thirds using spectral saliency in 9 quadrants. \\
  \bottomrule\\[-1ex]
\end{tabular}
\caption{Low-level aesthetic features extracted from each frame.}
\label{tab:computational-aesthetics}
\end{table*}

\subsection{Segment Features}

The proposed method computes segment-level features by aggregating frame-level features extracted from frames belonging to each segment.  Recall that each frame $\Frame$ is represented as a 62 dimensional feature $\Features{62}{\Frame}$.  Segment-level
feature for each segment $\Segment\In\Summary[\Video]$ is
\begin{equation}
\Features{124}{\Segment} =
\bigcup\limits_{i=0}^{61}
\left\{
  \Mean\left(\SetWhere{\Features{i}{\Frame}}{\Frame\In\Segment}\right),
  \Stddev\left(\SetWhere{\Features{i}{\Frame}}{\Frame\In\Segment}\right)
\right\}.
\end{equation}

\subsection{Ranking}\label{sec:segment-ranking}

We studied three models---(1) decision trees, (2) random forests, and
(3) XGBoost---for ranking segments using the segment features
discussed in the previous section.  We trained interestingness
prediction models for each of the above using segment-level features
extracted from videos available in \ac{SUMME} and \ac{TVSUM50}
datasets.  For training purposes these videos are divided into 5
second segments, and segment-level features are extracted for each
segment.  Train-test splits are generated using 10-fold
cross-validation on shuffled data, and the mean-squared-error is used
as the error metric for evaluating each model. The results for each
model are presented in \cref{tab:initial-model-training}.

\begin{table}[t]
\centering
\begin{tabular}{lllll}
  \toprule
  \textbf{Model} & \textbf{Min} & \textbf{Max} & \textbf{Mean} & \textbf{Std. Dev.} \\
  \midrule
  Decision Tree & $0.04005$ & $0.05145$ & $0.04559$ & $0.00380$ \\
  Random Forest & $0.02302$ & $0.03025$ & $0.02673$ & $0.00238$ \\
  XGBoost       & $0.02244$ & $0.02907$ & $0.02537$ & $0.00214$ \\
  \bottomrule\\[-1ex]
\end{tabular}
\caption{Mean-squared-error of each
  of our three base models evaluated using 10-fold cross
  validation. We can see that of the three models, XGBoost has the
  best performance, with the random forest model performing slightly
  worse, and the decision tree significantly worse.}
\label{tab:initial-model-training}
\end{table}

As we can see from \cref{tab:initial-model-training}, both the XGBoost
and random forest models obtain very similar error rates, with XGBoost
slightly out-performing the random forest model, and both
significantly out-performing the decision tree model. For this reason,
we will use both XGBoost and random forest models for evaluating our
system.

\subsection{Feature Importance}

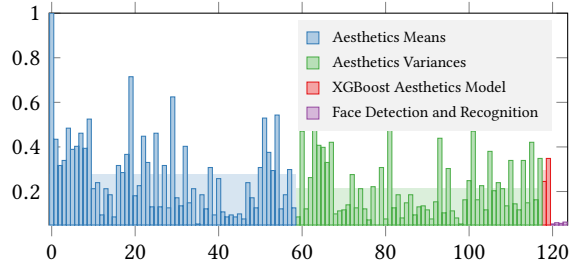
\begin{figure}
\centering
\begin{tikzpicture}[/pgfplots/ybar legend/.style={
    /pgfplots/legend image code/.code={%
       \draw[##1,/tikz/.cd,yshift=-0.25em]
        (0cm,0cm) rectangle (0.5em,0.5em);},
   }]
\begin{axis}[
  reverse legend,
  legend pos=north east,
  legend cell align={left},
  legend style={font=\scriptsize, draw=none, fill=black!5},
  ybar,
  width=\columnwidth,
  height=0.2\textheight,
  bar width=1,
  enlargelimits=false,
  enlarge x limits=0.005,
  bar shift=0pt,
  ]
\fill[plotd!20] (axis cs: 119.5, 0) rectangle (axis cs: 123.5, 0.058641653639048115);
\fill[plota!20] (axis cs:  117.5, 0) rectangle (axis cs: 119.5, 0.29677204562561516);
\fill[plotc!20] (axis cs:  58.5, 0) rectangle (axis cs: 117.5, 0.21259351180434727);
\fill[plotb!20] (axis cs:  -0.5, 0) rectangle (axis cs:  58.5, 0.2780892139169314);
\addplot[draw=plotd, fill=plotd!40] table [col sep=comma, x=feature, y=importance, select coords between index={120}{123}] {results/feature_importance_base.csv};
\addplot[draw=plota, fill=plota!40] table [col sep=comma, x=feature, y=importance, select coords between index={118}{119}] {results/feature_importance_base.csv};
\addplot[draw=plotc, fill=plotc!40] table [col sep=comma, x=feature, y=importance, select coords between index={59}{117}] {results/feature_importance_base.csv};
\addplot[draw=plotb, fill=plotb!40] table [col sep=comma, x=feature, y=importance, select coords between index={0}{58}] {results/feature_importance_base.csv};
\addlegendentry{\ \ Face Detection and Recognition}
\addlegendentry{\ \ XGBoost Aesthetics Model}
\addlegendentry{\ \ Aesthetics Variances}
\addlegendentry{\ \ Aesthetics Means}
\end{axis}
\end{tikzpicture}
\caption[Feature importances]{A plot of feature importances for each feature included in our final feature vector. For the purpose of visualization, we have grouped the features into four major groups, each represented by its own color; blue represents the mean values of aesthetic features, green the variances of theese aesthetic features, red the mean and variance of our XGBoost aesthetics model values, and finally purple the mean and variance values for our face detection and face recognition features. The background of each group additionally contains an aggregate bar which shows the average importance across the entire group.}
\label{fig:feature-importance}
\end{figure}

It is straightforward to compute feature importance when using
Decision Trees and XGBoost.  In order to see the efficacy of our
choice of features, we performed feature importance analysis.
Feature importance values are normalized between 0 and 1.  A value of
1 suggests that this feature plays an important role within the model.
Similarly, a value of 0 indicates that this feature is rarely used
during the prediction task.  \cref{fig:feature-importance} plots
feature importance for XGBoost model.

One important conclusion we can draw from
\cref{fig:feature-importance} is that among all the features used by
our model, face detection and recognition features have the least
average importance.  These features, incidently, are computationally
expensive to compute.  Our initial hypothesis was that the
computational cost of these features would be offset by their actual
importance when computing a segment
ranking. \cref{fig:feature-importance} shows that this is obviously
not the case.  We, therefore, decided to exclude face detection
and recognition features during segment ranking.  This leaves a 120 dimensional
feature for segment ranking:
$\SetWhere{\Features{120}{\Segment}}{\Segment\In\Segmentation[\Video]}$.

\cref{fig:feature-importance} suggests that features constructed using
XGBoost predictions have the highest average importance score.  Recall
that XGBoost model is trained on \ac{AVA} dataset.  This means that we
are able to train a supervised model for individual image aesthetics
and successfully apply this model to the task of segment ranking within
the context of video summarization.

\section{Video Summarization}\label{chap:summarization}

The final summary $\Summary[\Video]$ leverages segment rankings
$\SetWhere{\Ranking[\Segment]}{\Segment\In\Segmentation[\Video]}$
computed previously.  We formulate segment selection as a 0/1 knapsack
problem.  Given a set of items (segments)
$\Segment\In\Segmentation[\Video]$, each with a weight (duration) 
$\Frames{\Segment}$ and a value (ranking) $\Ranking[\Segment]$, we
determine which segments to include in our final summary such that the
final length is less than or equal to our target summary duration, and
the sum of segment rankings is maximized.  Mathematically, we can
describe this as
\begin{equation*}
\argmax_{\Summary \subseteq \Segmentation[\Video]} \sum_{\Segment\In\Summary} \Ranking[\Segment]
\Sep{subject to: }
\sum_{\Segment\In\Summary} \Frames{\Segment} \leq \TargetDuration.
\end{equation*}
This can be solved via dynamic programming \cite{knapsack}.
Define $T$ as an $n\times{}\TargetDuration$ array, and $T[i, \Weight]$ as the maximum score that can be obtained with duration up to or less than $\Weight$ using the first $i$ items of $\Segmentation[\Video] = \{\Segment[0],\ldots,\Segment[n-1]\}$.  We get the following recursive definition:
\begin{align*}
  T(0, \Weight) &= 0 \\
  T(i, \Weight) &=
            \begin{cases}
            T[i-1, \Weight] \\ \hspace{.5cm} \text{ if } \Frames{\Segment[i]} > \Weight \\
            \max(T[i-1, \Weight], T[i-1, \Weight - \Frames{\Segment[i]}] + \Ranking[{\Segment[i]}]) \\  \hspace{.5cm} \text{ if } \Frames{\Segment[i]} \leq \Weight. \\
            \end{cases}
\end{align*}
The solution can be found by computing the value of $T[n, \TargetDuration]$.

\section{Evaluation and Results}\label{sec:results}

We evaluate the proposed method using pairwise \FOneMeasure{} on
\ac{SUMME} dataset.  \ac{SUMME} contains multiple summaries from
different users, and we need a mechanism for comparing the summary
generated by our method with these user-generated summaries.
\cite{SUMME} proposed pairwise \FOneMeasure{} to perform this
comparison and evaluate the performance of a summarization scheme.
\FOneMeasure{} is computed as follows.  Given a summary 
$\Summary$ and a set of 
a set of user-generated summaries $\UserSummaries = \{
\Summary^0,\ldots,\Summary^n \}$, for each $\Summary^i$ in $\UserSummaries$ compute
\begin{equation*}
\Precision{i} = \frac{\Cardinality{\;\Frames{\Summary}\Intersect\Frames{\Summary^i}}}{\Cardinality{\;\Frames{\Summary^i}}}
\end{equation*}
\noindent and
\begin{equation*}
\Recall{i} = \frac{\Cardinality{\;\Frames{\Summary}\Intersect\Frames{\Summary^i}}}{\Cardinality{\;\Frames{\Summary}}}.
\end{equation*}
\noindent Pairwise \FOneMeasure{} is then
\begin{equation*}
\FOne{\Summary} = \frac{1}{n+1}\sum_{i=0}^{n} 2\cdot{} \frac{\Precision{i}\Recall{i}}{\Precision{i} + \Recall{i}}.
\end{equation*}

For the Random Forest and XGBoost models from
\cref{sec:segment-ranking}, we perform grid search over various model
parameters, and continue with the optimal parameters for each
variable. In the end, we compare the final pairwise \FOneMeasure{}
measures between the Random Forest and XGBoost models, and select the
model which attains the highest value.  Better methods are represented
by higher \FOneMeasure{} values.

Using the default parameters for our XGBoost model, our method
obtains an average \FOneMeasure{} value of $0.198$.  Average \FOneMeasure{}
scores obtained by competing methods in \cite{SUMME} and \cite{TVSUM} on
\ac{SUMME} dataset are $0.234$ and $0.2655$, respectively.  We fine-tuned
the XGBoost model for segment ranking.  The following parameters
were considered during grid search:
max depth, minimum child weight, gamma, subsample and col-sample
by-tree.  For our dataset, the optimal values for max depth,
minimum child weight, gamma, subsample and col-sample-by-tree are $3$, $5$, $0$, $1$ and $1$, respectively.
\FOneMeasure{} was improved from $0.198$ to $0.237$ using these
values.  

\begin{table*}[t]
\centering
\FMeasureTable{}
\caption[Accuracy Values for Various Methods on the \ac{SUMME}
  Dataset]{\FOneMeasure{} values resulting from testing various
  summarization methods on videos from \ac{SUMME} dataset. For each
  video, among the computational methods, the three highest results
  are highlighted using different shades of green. Darker shades are
  used for higher \FOneMeasure{} values, and hence better results.}
\label{tab:accuracy}
\end{table*}

\subsection{Accuracy on \ac{SUMME} Dataset}

We now compare our model to existing techniques on \ac{SUMME} dataset.
\cref{tab:accuracy} lists accuracy values for various methods on \ac{SUMME}
dataset.  Our method achieves the highest average \FOneMeasure{}
among the 5 computational video summarization schemes listed here.  
Average \FOneMeasure{} scores are provided for different videos in the \ac{SUMME}
dataset.  Our method posts the highest scores for roughly 50\% of the
tested videos.

\subsection{Performance}

\begin{table}
\centering
\PerformanceTable{}
\caption[Raw Performance Data for the \ac{SUMME} Dataset]{Raw performance data for our method applied to each video in the SumMe dataset. The duration of each video is provided, along with the time required for our method to complete, and corresponding speed as a multiplier of the duration of the video.}
\label{tab:performance}
\end{table}

\begin{figure}[t]
\centering
\begin{tikzpicture}
\begin{axis}[xlabel=Video Duration (s),
ylabel=Processing Time (s),
enlargelimits=false, width=0.9\columnwidth, height=0.2\textheight, legend style={at={(0.02, 0.96)},anchor=north west}]
\addplot table [only marks, x=duration, y=time, col sep=comma] {results/performance.csv};
\addplot [domain=0:400, color=blue] {0.614889692646 * x - 4.93780543364};
\addlegendentry{$y = 0.615 x - 4.94$}
\end{axis}
\end{tikzpicture}
\caption[Duration Versus Computation Time for the \ac{SUMME} Dataset]{A plot of the video duration versus computation time data from \cref{tab:performance}. We additionally plot a line of best fit to our data, demonstrating the fact that the complexity of our method appears to be linear in terms of the duration of a video.}
\label{fig:linear-performance}
\end{figure}
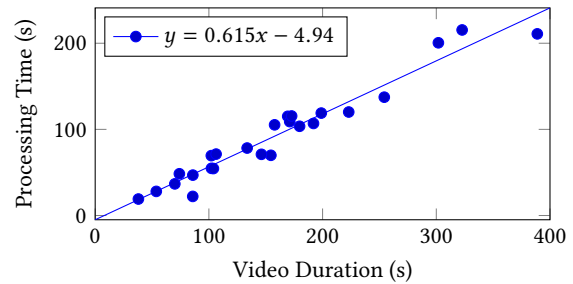

We performed video summarization for each video in the \ac{SUMME} dataset
using our method and recorded the times needed to generate the summaries.
These times are shown in \cref{tab:performance}.  Notice that summarization
times are smaller than the duration of the videos.  The third column
shows the speed of video summarization process.  On average our method
achieves a speed of $1.82$ times the actual duration of the video.  In
other words the time it takes to summarize a video is on average $0.55$
times the duration of the video.  \cref{fig:linear-performance} plots
summarization times vs. video durations.
It suggests a linear relationship
between summarization times and video durations.
We fit a first-degree polynomial to this data.  The coefficient of
determination for this fit is $R^2 = 0.943$, suggesting that
a line is a good estimator for this data.

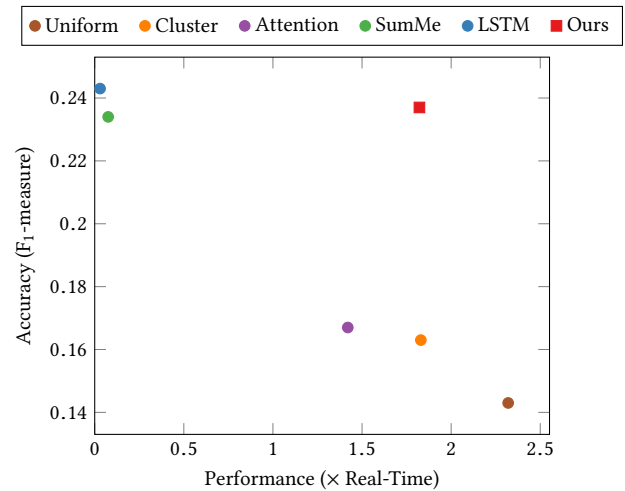
\begin{figure}[t]
\centering
\small
\begin{tikzpicture}
\begin{axis}[xlabel=Performance ($\times$ Real-Time), ylabel=Accuracy (\FOneMeasure),
legend style={at={(0.5, 1.03)},anchor=south,/tikz/every even column/.append style={column sep=0.75em}},legend columns=-1,
width=0.9\columnwidth, height=0.3\textheight,xmin=0,xtick={0,0.5,1,1.5,2,2.5}
]
 \addplot[color=plotg, only marks] coordinates { (2.32, 0.143) };
 \addlegendentry{Uniform}
\addplot[color=plote, only marks] coordinates { (1.83, 0.163) };
\addlegendentry{Cluster}
\addplot[color=plotd, only marks] coordinates { (1.42, 0.167) };
\addlegendentry{Attention}
\addplot[color=plotc, only marks] coordinates { (0.07547169811320754, 0.234) };
\addlegendentry{SumMe}
\addplot[color=plotb, only marks] coordinates { (0.03007518796992481, 0.243) };
\addlegendentry{LSTM}
\addplot[color=plota, only marks,mark=square*] coordinates { (1.82204501692, 0.237) };
\addlegendentry{Ours}
\end{axis}
\end{tikzpicture}
\caption{Average performance vs. accuracy.  Performance is the ratio of the video duration and summarization time.  A performance score of greater than 1.0 suggests that summerization times are less than video duration, i.e., it takes less time to summarize a video than it is to record this video. Higher performance valus are highly desireable.  Accuracy scores are average \FOneMeasure{}.  This plot also include performance and accuracy scores of a state-of-the-art LSTM-based method \cite{LSTM}.}
\label{fig:performance}
\end{figure}

\begin{figure}
  \centerline{
    \includegraphics[width=\linewidth]{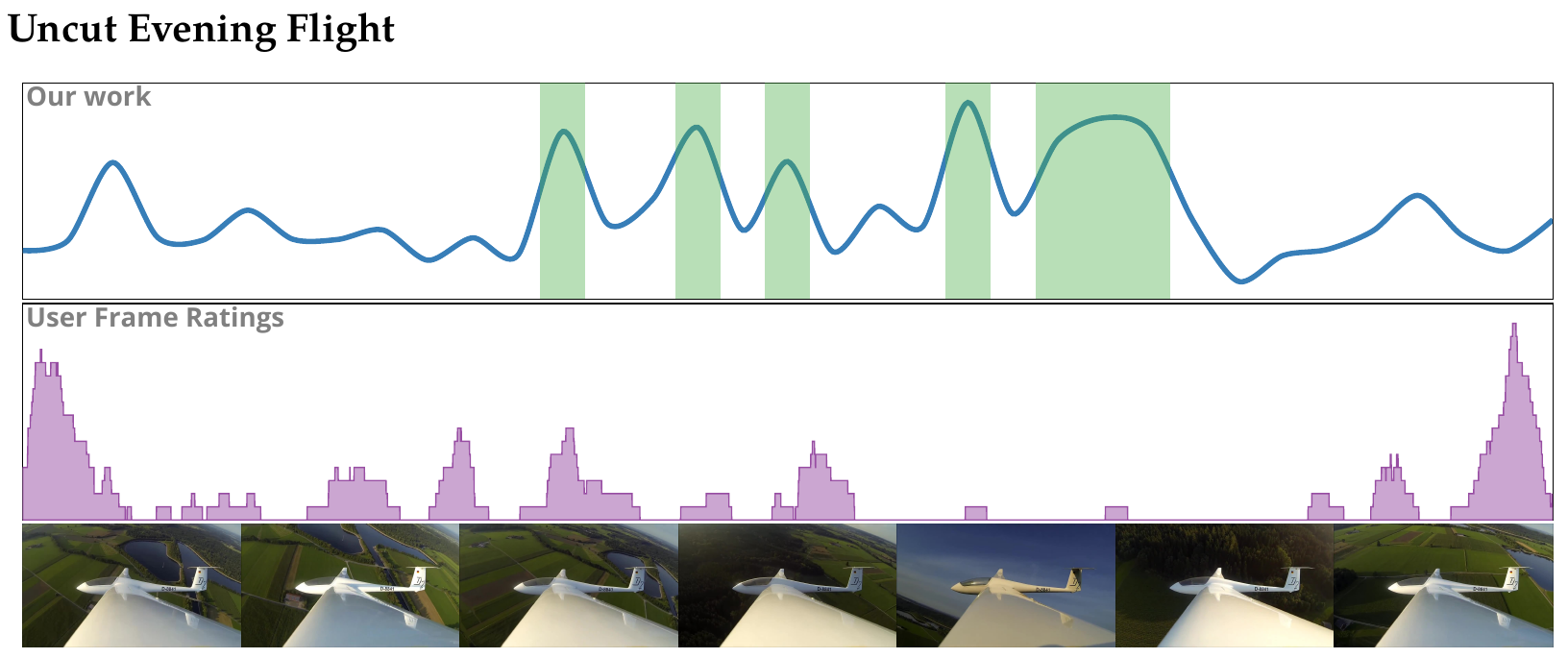}
  }
  \centerline{
    \includegraphics[width=\linewidth]{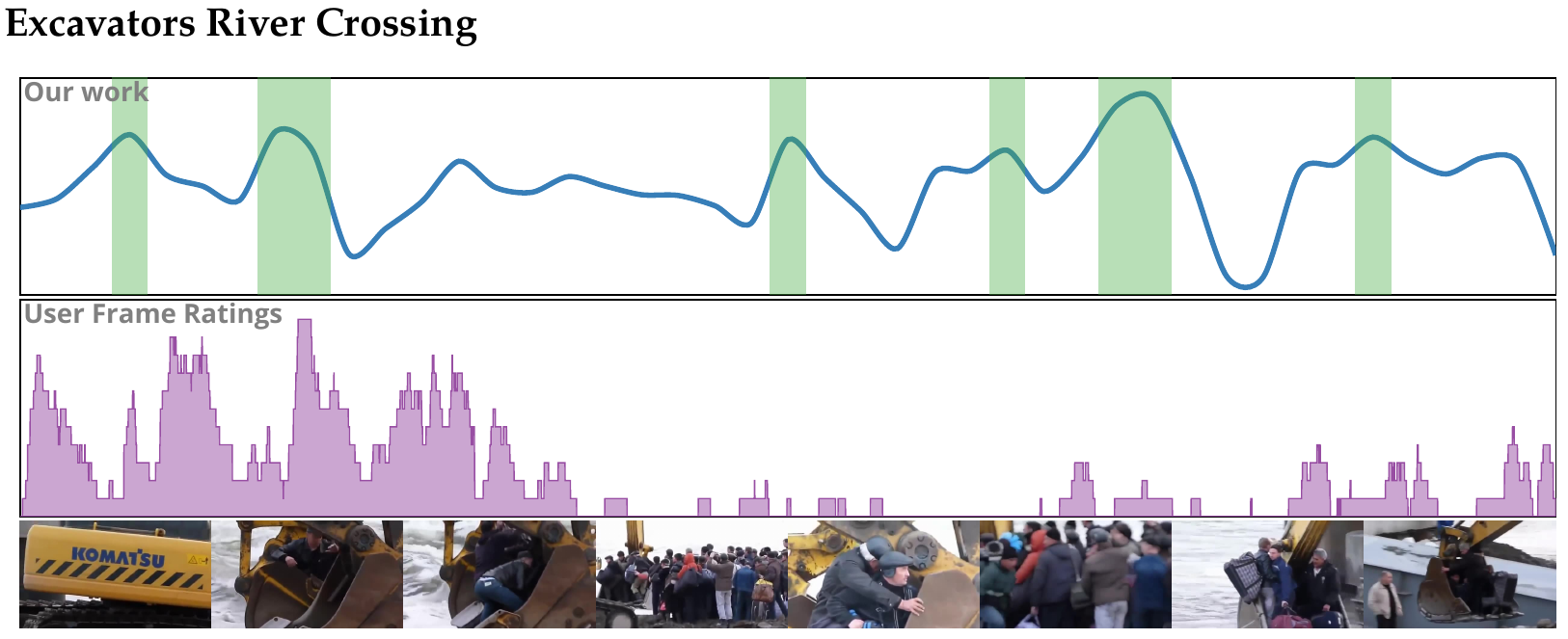}
  }
  \centerline{
    \includegraphics[width=\linewidth]{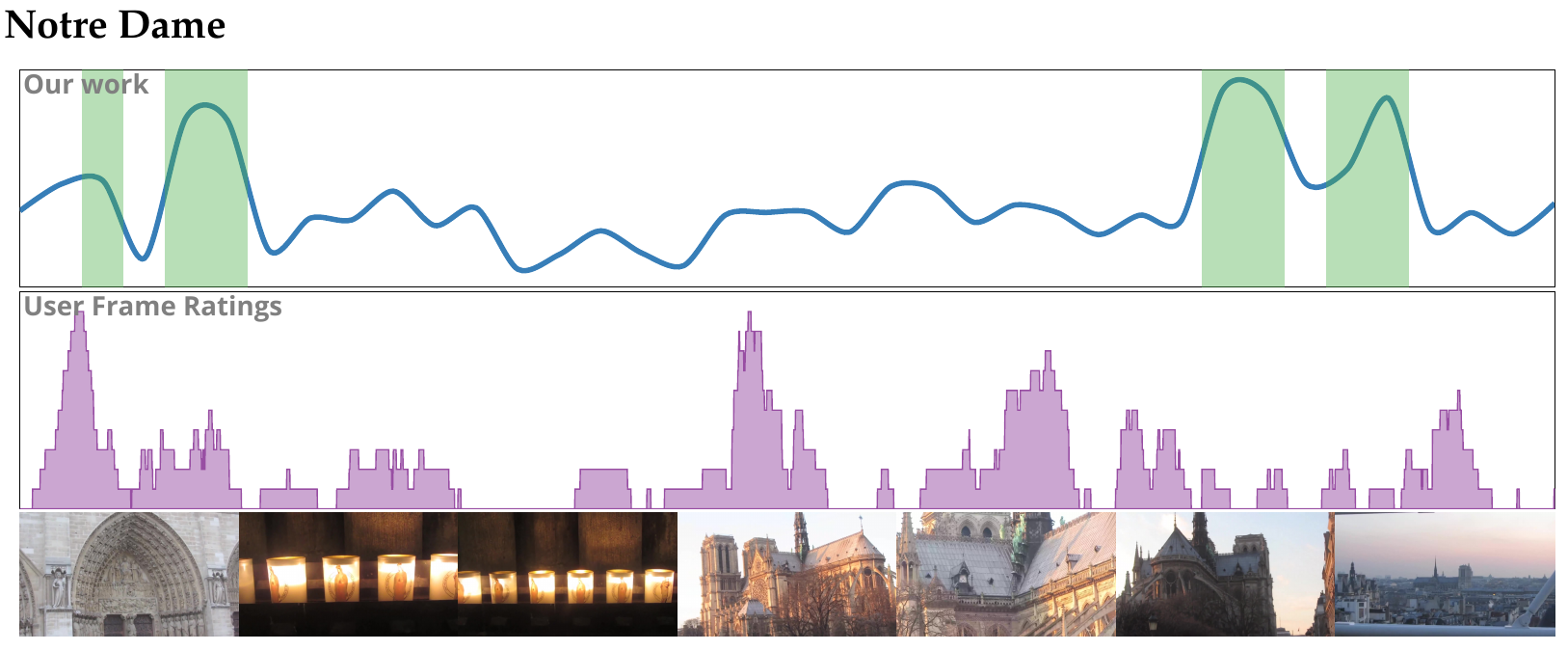}
  }
  \centerline{
    \includegraphics[width=\linewidth]{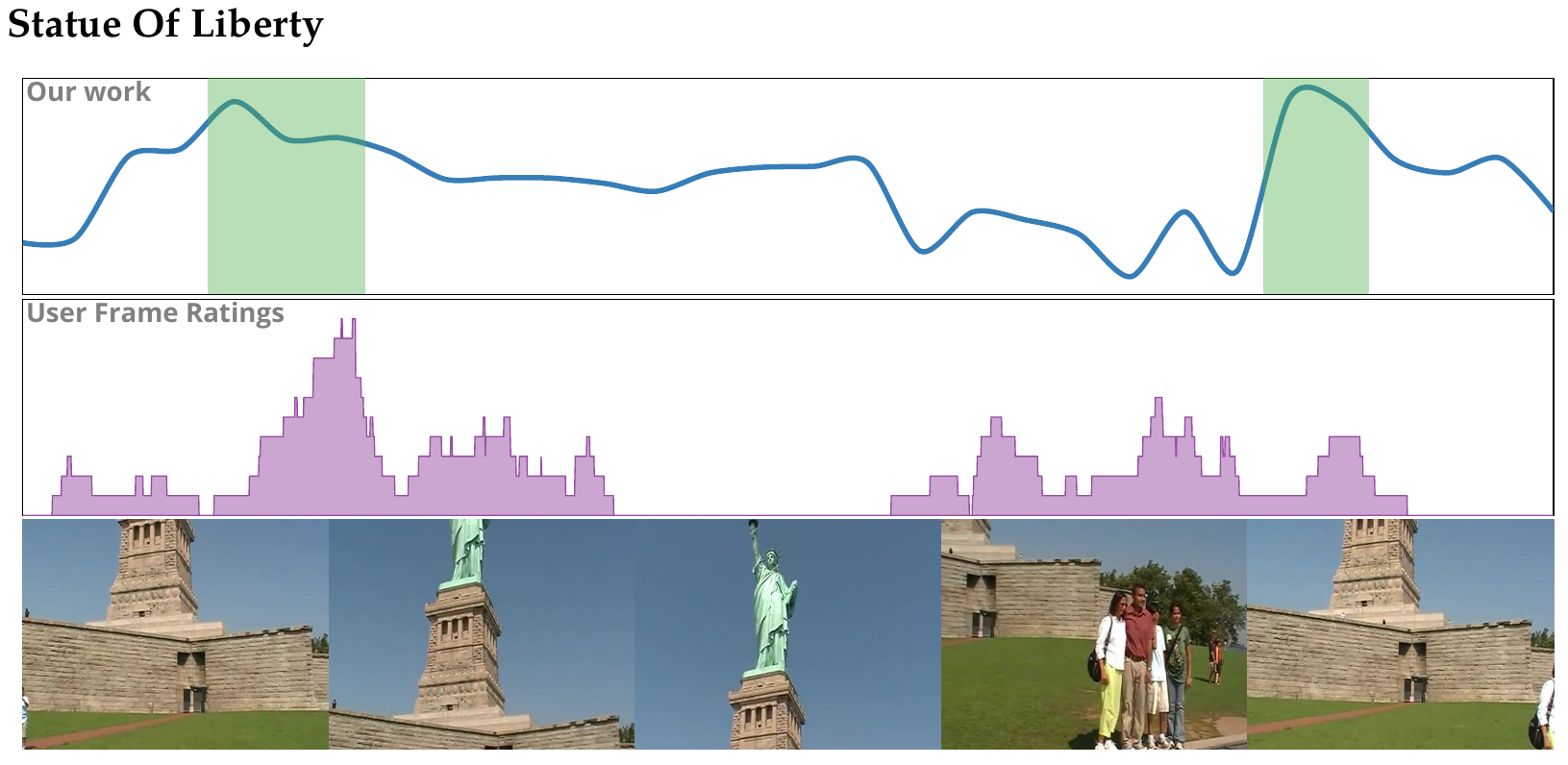}
  }
  \centerline{
    \includegraphics[width=\linewidth]{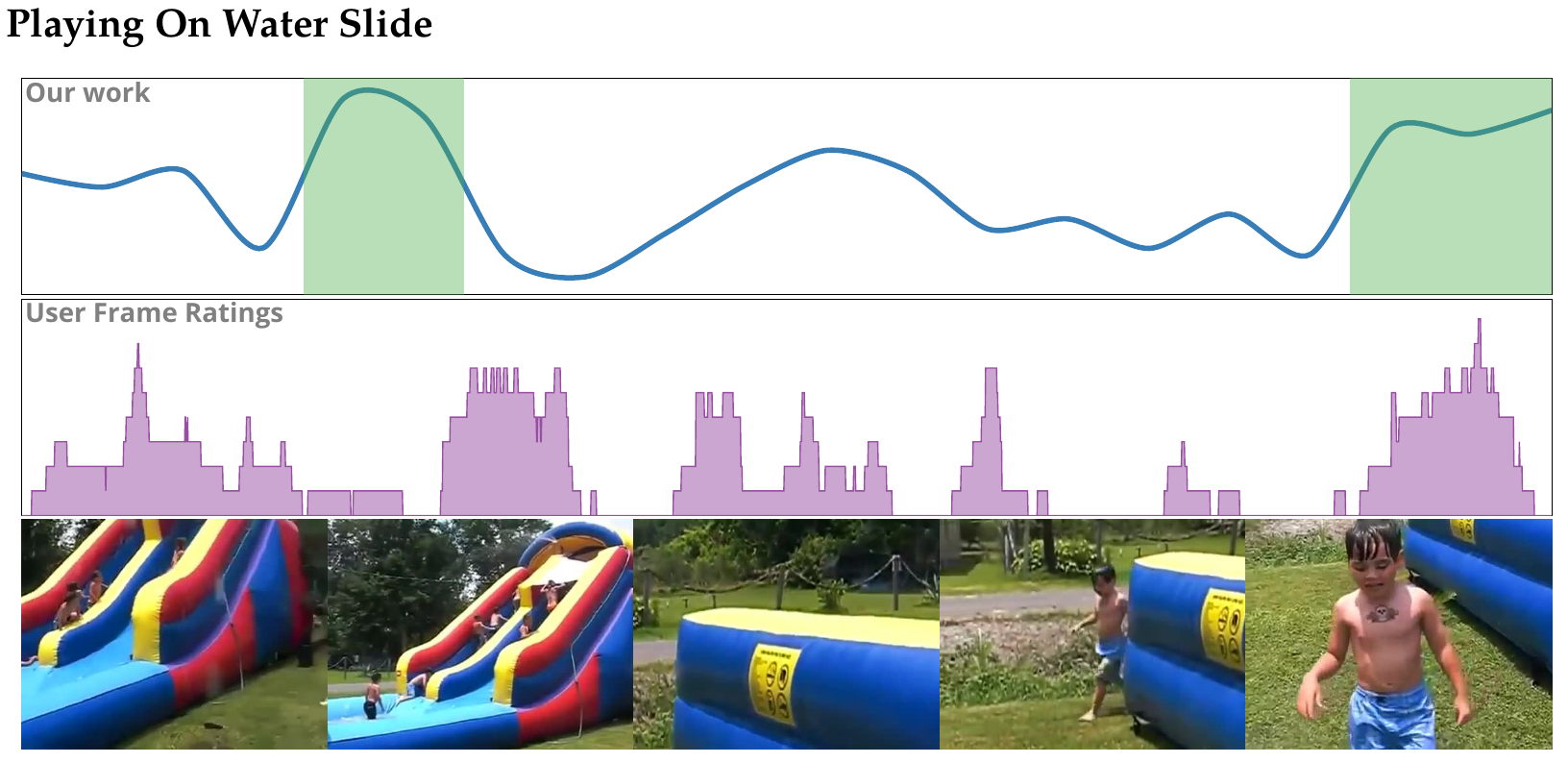}
  }
  \caption{Our method on a subset of \ac{SUMME} dataset.
    For each video, the top plot shows the segment ranking computed by
    the proposed method.  The middle plot shows the ground truth frame-level
    user rankings.  The last plot shows a selection of frames in the summary
  generated by the proposed method.}
\label{fig:examples}
\end{figure}

\cref{fig:performance} plots average performance vs. accuracy for different methods.  A performance value of 1.0 indicates that the summerization time is the same as the duration of the video.  We desire methods with performance greater than 1.0.  We can view these methods as {\it faster than real-time}.  Newer, computationally expensive methods---SumMe and LSTM---achieve high summarization accuracy; however, these methods posts poor performance.  Older, simpler methods on the other hand show high performance scores.  These methods, however, have low accuracy scores.  Our method is able to achieve high scores for both performance and accuracy.
Only the LSTM method is able to achieve a higher accuracy score than our method; however, the LSTM method has significantly lower average performance than our method.  \cref{fig:examples} shows summarization results for our method on a selection of videos taken from the \ac{SUMME} dataset.

\section{Conclusions}\label{sec:conclusions}

We propose a high performance video summarization system which is able
to perform video summarization in an online fashion on commodity
hardware.  The results demonstrate that our method is able to acquire
comparable summarization quality at a fraction of a computational
costs of a state-of-the-art LSTM method.  Our method, for example, is
able to create video summaries of arbitrary duration on a commodity
desktop---a \texttt{i5-3380M} CPU and with \texttt{16GB} of RAM and no
dedicated GPU---at times less than the duration of the videos.  This
suggests that our method may be ideally suited for mobile deployment.

The primary limitation of our method stems from how features are
computed for each segment.  We have chosen low-level features, which
are computationally inexpensive to extract.  A downside is that these
features are fundamentally limited in terms of capturing semantic
information present in a video.  We aim to solve this shortcoming in
the future by incorporating additional features into our framework.
We are also investigating methods to adapt our framework to
incorporate user preferences when creating video summaries.

\tiny

\end{document}